# Hotspot Prediction of Severe Traffic Accidents in the Federal District of Brazil

Vinicius Lima
Computer Graphics Technology/Criminalistics Institute
Purdue University/Federal District Civil Police
West Lafayette, IN, USA/Brasilia, DF, Brazil
Email: vlima@purdue.edu

Vetria Byrd
Computer Graphics Technology
Purdue University
West Lafayette, IN, USA
Email: vbyrd@purdue.edu

*Abstract*—Traffic accidents are one of the biggest challenges in a society where commuting is so important. What triggers an accident can be dependent on several subjective parameters and varies within each region, city, or country. In the same way, it is important to understand those parameters in order to provide a knowledge basis to support decisions regarding future cases prevention. The literature presents several works where machine learning algorithms are used for prediction of accidents or severity of accidents, in which city-level datasets were used as evaluation studies. This work attempts to add to the diversity of research, by focusing mainly on concentration of accidents and how machine learning can be used to predict hotspots. This approach demonstrated to be a useful technique for authorities to understand nuances of accident concentration behavior. For the first time, data from the Federal District of Brazil collected from forensic traffic accident analysts were used and combined with data from local weather conditions to predict hotspots of collisions. Out of the five algorithms we considered, two had good performance: Multi-layer Perceptron and Random Forest, with the latter being the best one at 98% accuracy. As a result, we identify that weather parameters are not as important as the accident location, demonstrating that local intervention is important to reduce the number of accidents.

*Keywords-traffic accident; machine learning; forensic sciences; accident reconstruction; hotspots.*

## I. INTRODUCTION

Traffic accidents are a pervasive problem in modern cities worldwide. In response, the World Health Organization (WHO) has created an ambitious agenda to reduce accidents by 2030 [1]. In Brazil, traffic accidents are a major public health concern, with the government spending significant amounts of money on addressing their consequences [2]. However, instead of reacting to accidents after they happen, prevention is key. Machine Learning (ML) [3] is a rapidly advancing technology that can predict outcomes based on features and provide valuable insights into their causes. These methods can be useful tools for accident analysis and align with the concepts of Safe Smart Cities [4].

The increasing number of traffic accidents creates an opportunity to study them and get a better perspective on their causes. The knowledge derived from the reason behind traffic accidents can help authorities make better decisions regarding public policies or infrastructure to prevent future cases.

This paper focuses on data collected by crash reconstructionists from the Brazilian Federal District Civil Police. In Brazil, every accident with injuries requires a forensic team to collect evidence and analyze the scene, providing a forensic report to police investigators and justice authorities [5, p. 9]. Since 2019, the forensic team has been collecting data in a digital format, creating an archive of accident data and information. In practice, accidents that resulted in minor injuries or the ones that could not be isolated by the police (for safety reasons, or to relieve traffic congestion, for instance) are not examined on-site, but instead, they have the vehicles removed for analysis at a police station. Here, we define severe accidents as the ones where at least one person was taken to the hospital or died.

The motivation for this research came from a previous visualization tool that identified three hotspots and analyzed their causes based on forensic reports [6]. This work demonstrated the importance of studying crash concentrations and their causes. Forensic analysts typically focus on individual cases, but this paper aligns with the concept of "Forensic Intelligence" by exploring the broader law enforcement knowledge that can be gained from analyzing the rich database collected from many accidents [7]. Therefore, we demonstrate the opportunity forensic analysts have to provide intelligent data by studying the general behavior of accidents, abstracting hidden patterns, and consequently creating supportive knowledge to prevent new cases.

A prediction analysis was conducted based on the data collected on-site in conjunction with weather parameters in the Federal District of Brazil, where the capital of the country sits. ML algorithms such as Random Forest and Multi-layer Perceptron, demonstrated to have good performance to predict the number of accidents in different regions, therefore, identifying hotspots of accidents. Furthermore, a feature importance analysis based on the evaluated parameters was conducted and described.

The remainder of this paper is structured as follows. Section II provides an overview of similar and related work. Section III summarizes the primary objectives of our research, outlining the research questions we aim to address. In Section IV, we present the dataset used for this study and detail the preprocessing steps undertaken to ensure its suitability for our objectives. Section V explains the methodology employed in this project, while Section VI provides a comprehensive discussion of the main findings. Additionally, in Section VI, we acknowledge the limitations of this work and outline our plans for future research. Finally, the article concludes with Section VII, where we present our conclusions, followed by an acknowledgement section.

## II. Related Work

The study of traffic accident causes is not new, as well as the prediction of accident hotspots. Nevertheless, the literature is diverse in the way the analysis is conducted. Authors usually use case studies to state their points of conclusion [8]. Identifying the main reasons behind an accident is hard, subjective, and likely to be dependent on several features. The heterogeneity of the results is corroborated by the research of many authors exploring the subject in different datasets from different angles. We will focus on related literature that uses ML in traffic accident datasets on a city level. Also, we are interested in previous work that aimed to detect hotspots of accidents. It was not part of our review literature scope, papers that address accident prediction for autonomous vehicles purposes.

Regarding the related literature that uses machine learning as a supporting technique to predict future accidents, they mainly follow two approaches [9-10]. The first one is the prediction of the accident itself. They mainly follow a simple supervised approach of using a dataset to be trained, where the output is binary classification into accident or not-accident. However, the data used only have information about when the accident happened (positive samples). Therefore, a common technique required is to generate negative random samples in order to have information on non-accident cases [11-15]. This is not a trivial task, and its foundation is somewhat controversial. Most of the studies generate negative samples in an arbitrary way, which can be questioned. They usually have labeled non-accident data as an expansion of the real accident information. For instance, Hébert, et al. [12] present that for each accident, there are three non-accidents, based on the combinations of time and day for non-positive cases. Although there is a reasonable explanation for the generation of negative samples, it is unsure if this technique can be expanded for every city and any dataset. To avoid this issue, other works focus not on the prediction, but on the level of severity of an event [16-18]. These are multiclass classification problems, where the output goal is a scale of how severe an accident can be. Nevertheless, the results of this approach are important for public policies, someone can argue that severity can be quite subjective, as the definition of severity may vary within regions. In contrast, this research provided an analysis using only objective data, without creating fictional information, by targeting the number of accidents in an area.

As for the prediction of hotspots, we have less research. Here the outcomes are treated as a regression problem, instead of a classification. Lu et al. used Linear Regression to check whether certain regions of Beijing were considered hotspots of accidents based on data from the road, environment, vehicle, and driver information [19]. Mansourkhaki et al. presented a hybrid model of accident prediction combining a prediction analysis with statistics functions and previous knowledge to filter out outliers in big sets of accident data [20]. Ren et al. used the deep learning technique of Long Short-Term Memory (LSTM) to predict traffic accident risk by analyzing the Spatio-temporal characteristics of accidents in Beijing [21]. Yuan et al. created a similar analysis but predicted the number of accidents in real-time by training the past seven days of data from Iowa, using ConvLSTM (Convolutional Long Short-Term Memory) [22]. Fawcett et al. presented the prediction of the number of accidents in the city of Halle (Germany) using the Bayesian approach [23]. Al-Omari et al. used GIS and Fuzzy Logic to identify hotspots in Irbid City, Jordan [24]. Lin et al. proposed the prediction of real-time accidents based on the FP-tree (Frequent Pattern tree) on Virginia's I-64 highway [25]. Here again, we see how diverse the research in the field is, with different algorithms providing solutions tailored to the studied city and dataset.

As for a similar work regarding data from Brazil, only one case was found using machine learning to study the causes of road accidents on a highway south of the country [26]. To the best of our knowledge, this is the first time a prediction analysis is conducted from Brazilian data at a city level with data from traffic reconstruction analysis.

Our goal is to use ML to predict the number of accidents, based on data from crash scenes and weather data. We aim to contribute to the diverse literature, with a novel dataset and location, adding more knowledge to the study of causes behind vehicle accidents. More importantly, we want to provide local authorities with supporting information on how ML can be used to help decisions to prevent new collisions in future intelligent cities. Finally, it is also our goal to demonstrate how crash forensics teams can use their data to extract important high-level information, expanding their work to a collective perspective, instead of just individual investigations.

## III. Research Design

To achieve high steps of insights it is interesting to study the intersection between distinct domains or datasets [27]. In our research, we do this, by connecting our accident data with weather information and trying to check possible correlations.

We are mostly interested in the following research questions:

1. Is it possible to the predict number of accidents in defined regions (i.e., hotspots) with traffic accidents and weather information?
2. If yes, what are the main features involved in the prediction outcomes?

We will first present our datasets, give an overview of the preprocessing phase, and a brief analysis of heatmap visualizations that are more useful for our purposes. Then we will proceed to the ML algorithms, with the resulting performances. Then, we will provide an analysis of the correlation between the predictions and the features involved. Finally, we will discuss the results and how the current work can aid government authorities in strategies to prevent future accidents.

## IV. Dataset and Prepossessing

Two datasets were used for this research. The first one was the severe accident data originating from the Federal District Civil Police (*Polícia Civil do Distrito Federal*), regional police situated in Brasilia, the capital of Brazil. The data was provided by a non-profit organization that fosters research in forensic sciences, *Fundação de Peritos em Criminalística*

*Ilaraine Acácio Arce*, which is formed by forensic scientist officers. It contained information from the accident scene, such as the topography of the street, road conditions, vehicle information, and damages, collected in the years 2020 and 2021. The total number of severe accidents presented was 3,846, with almost half each year. For the purpose of this work, we focused on objective data only, i.e., numbered information such as GPS coordinates (latitude, longitude), date, time, and road speed limit. The second dataset is the weather information, which came from the National Meteorology Institute (Instituto Nacional de Meteorologia). It contained instant values of temperature, humidity, pressure, wind, solar radiation, and rain precipitation from five different stations in the Federal District.

The first step was to preprocess the data in order to filter out null information and mistaken outliers (when the coordinates were situated outside the limits of the Federal District, for instance). Then, we needed to merge the accident dataset with the weather into one. The weather information came from five stations spread out in the district. To this objective, it was necessary to match each accident with information from the closest meteorologic station, according to the time it happened. For each accident location, we measure the distance from each of the five stations and merge the weather information with the closest one. Our weather dataset provided hourly measurements, so we matched the hour of the accident with the hour of the collected meteorology information (this generalization is understandable, since in a tropical country like Brazil, the temperature has low variance within the hour). One important consideration is that our accident data does not provide the time when the accident actually happened, but only when the forensic team got to the scene. However, in previous research, it was possible to generalize that on average the forensic team gets to the scene about one hour after the accident really happened [6].

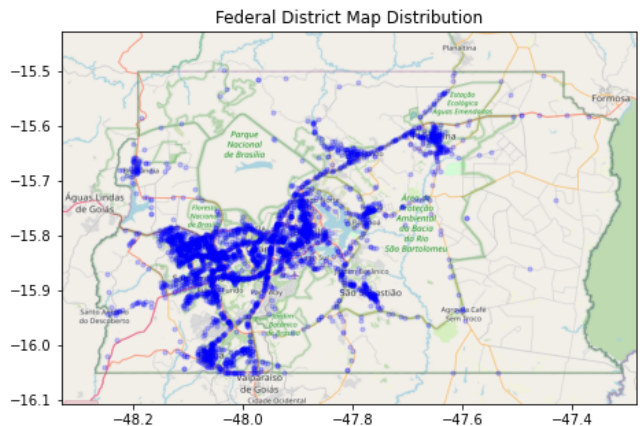

Figure 1. Scatter plot of accidents in 2020 and 2021. It can be seen that there is a contraction of accidents in urban areas. Python and Matplotlib were used to generate this map and the axes represent latitude and longitude coordinates.

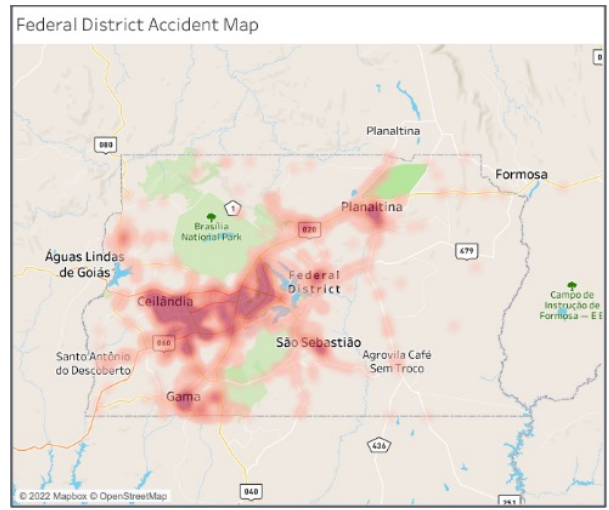

Figure 2. Density map of accidents in 2020 and 2021. The darker color represents hotspot areas of crashes. Tableau was used to generate this map.

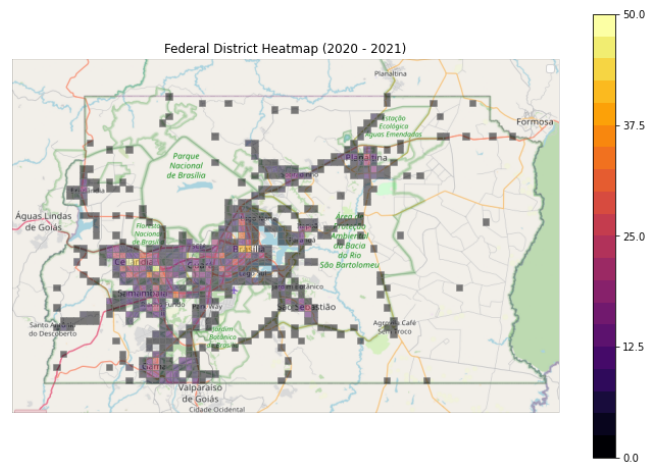

Figure 3. Density gridded map of accidents in 2020 and 2021. Colors close to yellow correspond to higher numbers of accidents. Python and Geopandas were used to generate this map.

Figures 1 and 2 represent the location of accidents as a scatter plot and density map, respectively. We want to predict the number of accidents in defined regions; therefore, it is important to divide our map into grids and count the number of accidents in each grid. Figure 3 shows this representation, using a colormap to visualize "hotcells" based on the number of accidents in each grid. The map was divided into 80 regions, which gives us a grid with dimensions of 4.6 miles by 6.0 miles. This representation is more likely to be used by the police or authorities to identify problematic regions since it provides exact boundaries of these regions. While scatter plots and density maps also provide some insights into the concentration of events, it is hard to accurately identify the limits of each hotspot location. Moreover, the width and length of the grids can be set accordingly to how authorities want to look at the problem. For instance, they can set the cells at a road intersection level or a neighborhood level.

## V. METHODOLOGY

Machine Learning algorithms are mostly used to solve regression or classification problems. Since we are trying to predict the number of accidents, we are dealing more with a regression situation, but we tested both approaches with the main algorithms. As it will be detailed later, Random Forest, Multi-layer Perceptron demonstrated to be an efficient model for the task in question. Other algorithms, such as Support Vector Machine [28], Linear Regression [29], and Neural Networks [30] were also tested, but performed poorly or took too long to provide the prediction. To keep it short, we will not describe their details.

As mentioned, we combined information from traffic accident data and climate data, having the following inputs: Longitude, Latitude, Street Speed limit, Year, Month, Workday, Hour of the day, Temperature, Humidity, Pressure, Wind, Solar Radiation, Rain Precipitation. The output of the model was the number of accidents in boundary regions. We set these regions, divided our map into grids, and counted the number of accidents in each cell. For each accident, we added this value as a target prediction. We trained our model with data from 2020 and tested it on the 2021 dataset, in order to check the correlations between consequent years. We used Python along with the Scikit-learn library to implement the models.

After preprocessing the data, it is time to build the algorithm to be trained. Basically, what these techniques are trying to do is solve an optimum way to classify (or reduce) the output based on the given input information. Random Forest achieves the solution by creating several decision trees and averaging the results of each ramification. Multi-layer Perceptron Classification does the same task by having an artificial network of neurons that sets different weights to the nodes that closely fit the output. For this technique, we used 3 hidden layers, with 21 neurons each, and the ReLu (Rectified Linear Unit) default as the activation function. Performance and results are discussed in the next section.

## VI. RESULTS AND DISCUSSION

Table I provides the accuracy of the tested algorithms, both treating the problem as regression or classification.

TABLE I. ALGORITHMS COMPARISON

| Model | Type | Accuracy |
|---|---|---|
| Random Forest | Regression | 98.8 % |
| Multilayer Perceptron | Regression | 79.6 % |
| Random Forest | Classification | 94.1 % |
| Multilayer Perceptron | Classification | 88.2 % |

As can be observed, Random Forest performed really well in both situations, achieving almost 99% of the right predictions. This can be explained by looking more deeply into the data and observing that the locations of accidents do not differ much from one year to another. This information already reveals that from 2020 to 2021 not much was done to change the traffic scenario.

The good performance of both models can be explained by their capability to reveal a pattern behavior with non-linear features. However, as we increase the number of grids, the accuracy decreases. This makes sense since with more regions on the map, it would be harder for the model to correctly predict the number of accidents. Moreover, we saw a greater decrease in the accuracy of the classification models when the number of grids raises. This is somewhat expected since the classification model treats the number of accidents as categories, making it hard to predict for a wider range of labels. The classification would be suitable for authorities to predict levels of accident concentration (for instance, on a scale of low to high). The developed model has been demonstrated to be efficient for a certain number of grids in the map and thus could be easily adapted to fit the prediction into hotspot categories. The regression approach treats the values as a continuum parameter, and it is suitable in case authorities want to predict the actual number of accidents in a given area. This technique has demonstrated good results even with a different set of grids. However, more data is important to give stronger support to decision-makers.

Figure 4 shows the features extracted from the Random Forest technique in terms of importance to the model's output. It can be seen that the location attributes (latitude and longitude) are the main features contributing to the prediction. We can deduct from the figure that weather conditions do not influence severe accidents in a significant way. When taking out latitude and longitude from our model, Random Forest returns a poor accuracy of only 0.04%. Even the pressure attribute, a weather parameter, can be said to be related to location since this value changes according to the altitude level.

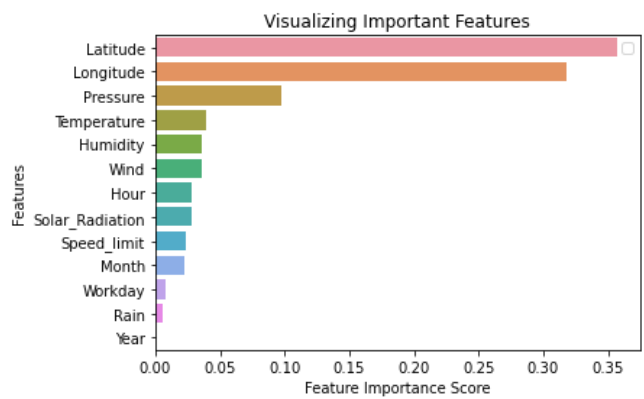

Figure 4. Density Histogram of ranked features used in the Random Forest algorithm. To each feature, it is assigned the proportion of which each contributes to the outputs.

The non-correlation with weather conditions actually aligns with related research [31-32]. This may be explained because the weather in tropical places is almost the same throughout the year and the same result may not apply to regions where the seasons are more well-defined. Another interesting result is that the hour and month of the accident are also less important features when predicting the number of accidents. This infers that rush hours may be an important factor to predict hotspots, but not as much as the location of the accident itself.

Based on our findings, we can conclude that if we want to understand more about the causes of the accidents, we need to study deeper the characteristics of the location of the accident rather than looking at other variables, such as the climate conditions. In the end, hotspots' identification and understanding are key factors to act in the direction of traffic accident prevention.

## VII. Limitations

For the purposes of this research, we used all the data available from traffic reconstructionists. However, it is important to state that our dataset is considered small for prediction purposes. Only about 1,600 points were used to train the models. This also explains the high accuracy result. More data means more information, consequently providing more realistic outputs, and possibly a lower performance is expected. This research worked as a proof of concept to the claim of automatically predicting hotspots of accidents, providing a powerful tool to support knowledge regarding accident prevention. Moreover, as mentioned before, the number of grids can be set, according to how big the authorities need the region to be. Naturally, the accuracy of the results tends to decrease as the number of grids rises (another issue that can be resolved by having a bigger dataset).

Also, as mentioned in the Introduction, data used to feed the algorithms were from severe accident cases that had an on-scene examination. Thus, it does not contemplate the overall traffic accident cases.

## VIII. Conclusion and Future Work

The present work is a case study of a real-world dataset about severe traffic accidents collected from the Federal District Police in Brazil. The data were analyzed along with weather information such as temperature and rain precipitation. After prepossessing the data, we first visualized some of its content to get a sense of the accident panorama. We were particularly interested in predicting the number of accidents in defined regions and, thus, identifying intelligent hotspots based on past evidence. For this work, we tested traditional machine learning algorithms, analyzing the performance as regression and classification problems. Random Forest and Multilayer Perceptron outputted high accuracies, with the former outperforming the latter.

Random Forest model also revealed that location attributes, such as longitude and latitude, were considered more important features to predict the number of accidents than the meteorology parameters, such as rain and temperature, and time parameters, such as hour and day. In other words, the place where the accident happened is a key factor contributing to the identification of hotspots. Public authorities and researchers need thus to focus their studies primarily on the site characteristics, for example, suggesting road infrastructure interventions. Overall, we demonstrate how density maps and machine learning are powerful tools to generate insights for high-level decisions in the direction of accident prevention.

To the best of our knowledge, this is the first work with machine learning in traffic accidents in the Federal District of Brazil. Thus, we add new insights to this diverse research area, which attempts to understand the root causes of vehicle collisions. Moreover, we demonstrated that data from forensic teams can be used not only for traditional one-by-one investigations but also to understand patterns and associations in the big picture, delivering useful information for public policies. This work aligns with the new era of Forensic Intelligence, where data science plays a key role.

For future research, beyond more points, we will add to the models discrete variables collected from the traffic reconstruction forensic team. As a result of this work, it will be interesting to observe the performance considering location attributes, such as topography, cross intersections, junctions, street lighting, and road conditions.

## Acknowledgment

We appreciate the support of the non-profit organization *Ilaraine Acácio Arce* and the Federal District Civil Police, in Brasilia, Brazil, for providing data and investing in research.